**Graphcut Texture Synthesis for Single-Image Superresolution**

Douglas Summers Stay

Master's Thesis

Graduate School of Arts and Sciences

New York University

2006

# Graphcut Texture Synthesis for Single-Image Superresolution

**Abstract:**

Texture synthesis has proven successful at imitating a wide variety of textures. Adding additional constraints (in the form of a low-resolution version of the texture to be synthesized) makes it possible to use texture synthesis methods for texture superresolution.

## The Single-image Superresolution Problem

The problem we are trying to solve is the following:

Given:

a low-resolution image and

high-resolution versions of each texture within the image

create:

the corresponding high resolution image

The basic assumption behind *texture superresolution* is the following: if we build a copy of the low-resolution image using reduced resolution patches from sample textures, placing the corresponding high-resolution patches in the same positions will create a plausible reconstruction of the missing high-resolution image[1].

There are two main obstacles to getting high quality results. First, there needs to be enough of the sample textures to be able to find a good match for each low-resolution patch. Second, the seams between neighboring patches must be created in such a way as to minimize the transition while retaining detail.

---

[1] There is a scene in the TV series "Kung Fu" where the young acolyte Caine is assembling a puzzle of a map of the world. He discovers that if the pieces are flipped over, they can be assembled into the picture of a man. Since Caine knows what a man is supposed to look like, this is easy to do, and the result is that the map of the world is assembled without ever having looked at it. In the same way, we assemble low resolution patches into a copy of the low resolution image that we have, in order to place high resolution patches to create an image that we don't know yet.

The main difference between this program and previous patch-based super resolution programs is its focus on reconstructing textures well. (In [15], for example, they use similar sampling methods but explicitly avoid reconstructing texture entirely.) Although these other programs create a visually pleasing reconstruction, they tend to err on the side of caution, not adding any information unless there is a clear signal to support it. This program, on the other hand, is designed to retain as much of the detail present in the source patches as possible.

There are multiple reasons for focusing on reconstructing textures accurately:

1.) Other sophisticated interpolation techniques (such as NEDI or Jensen's approach) perform well on smooth regions, edges, and corners, but create unsatisfactory artifacts in areas of texture. A common failure is oversmoothing. This is understandable, since smoothness is a good general assumption when it comes to images. But we want to learn what assumptions to make for each particular texture rather than simply use a smooth prior.

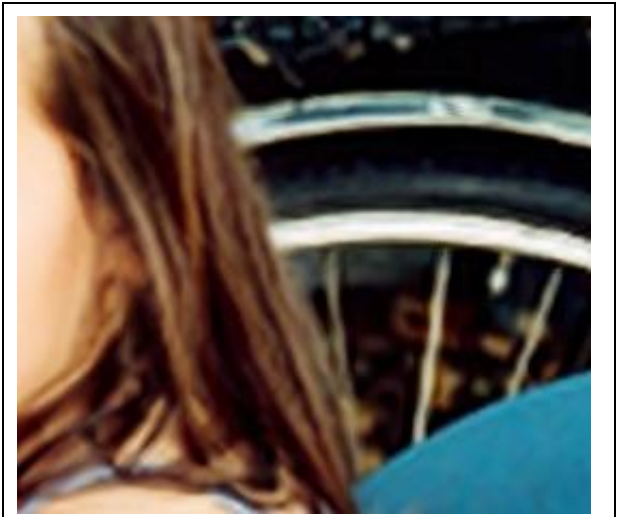

New Edge-Directed Interpolation performs well on the bicycle wheel, which has smooth areas and sharp edges. Areas of strong texture (such as the hair) look as if they have been painted on or as if they are made of plastic, due to an assumption of smoothness.

2.) Trying to reconstruct a texture is easier than trying to reconstruct a natural image. The chance of finding an accurate match is high because a texture by definition contains many similar regions at slightly different scales, orientations, and shapes.

3.) Even if the reconstruction is incorrect, it will still often be plausible because it comes from the correct distribution. When enlarging a picture of a forest, it doesn't matter for most applications that all the leaves are in exactly the right place, as long as the distribution of leaves looks realistic and the leaves are in generally the right areas. (This philosophy makes it difficult to quantitatively measure the quality of a reconstruction, however: common image error metrics like PSNR would find two images of pure noise to be maximally different, while to the human eye they would be effectively identical.)

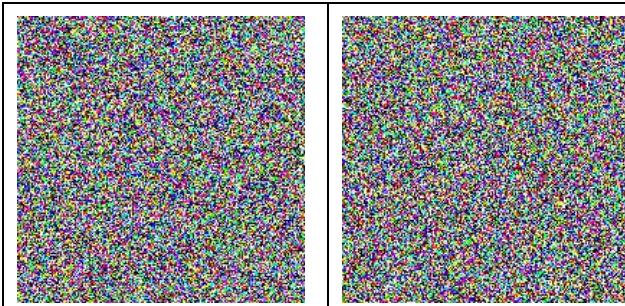

These two images of pure nosie appear to be essentially the same, though at a pixel level the colors are completely different. If we were to replace one by the other, the change would be unlikely to be noticed. This program makes use of this fact to reconstruct textures in a way that is acceptable to the human visual system.

Texture synthesis and texture matching are well-explored areas in graphics, with hundreds of papers having been written about each of them. This program takes ideas from among the most successful texture techniques (graphcut texture synthesis, and combined histogram and local feature based texture matching) to improve on previous results.

**Outline of program structure**

In the following outline, $E(x, F)$ is the operation of enlarging by a factor F using bicubic interpolation (or any other interpolation method). $E^{-1}(x, F)$ means reducing by the same factor. $B(x)$ means perform a convolution (a blur) with gaussian kernel of radius $F$. The symbol $\otimes$ is used to distinguish patches and images that are band-pass filtered.

1.) Load two images: the *target image* $T$ to be enlarged, and the *source image* $S$ containing the high frequency details to be learned. (Rotations or other transformations of the source images can also be searched.)

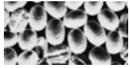
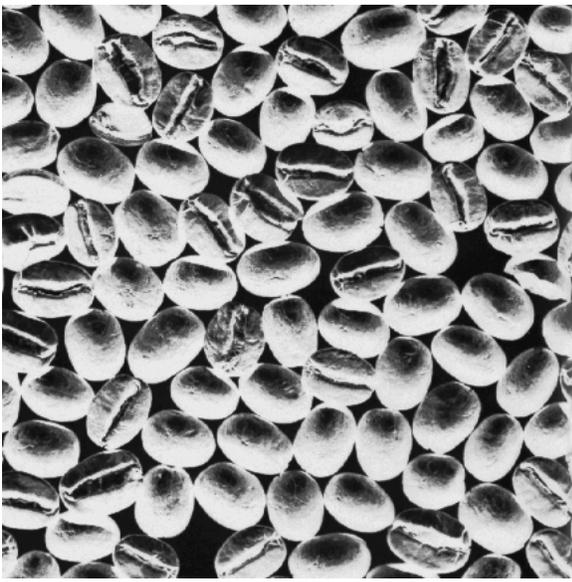

Target $T$    Source $S$

2.) Enlarge the target image by the *enlargement factor* $F$ using a simple interpolation scheme, such as bicubic interpolation. Call this the *enlarged target* $E(T,F)$.

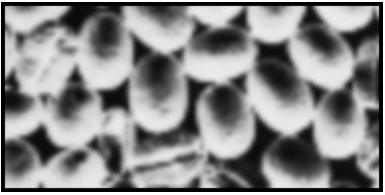
enlarged target $E(T,F)$

3.) Reduce and re-enlarge the source images by the enlargement factor, using the same interpolation scheme (bicubic, usually). Call these the *reenlarged source* $E(E^{-1}(S,F))$.

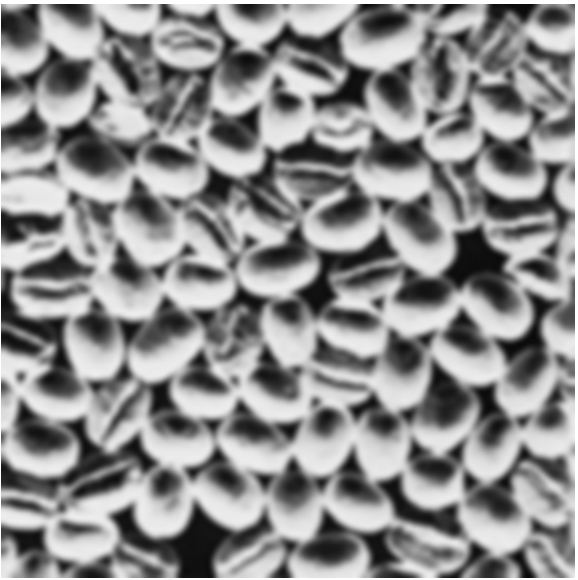
reenlarged source $E(E^{-1}(S,F))$

4.) Subtract a blurred image of the enlarged target from itself
$T\otimes = E(T,F) - B(E(T,F))$, and a blurred version of the reenlarged source from itself
$S\otimes = E(E^{-1}(S,F)) - B(E(E^{-1}(S,F)))$. This leaves only the medium frequency information in the enlarged target and reenlarged source.

| 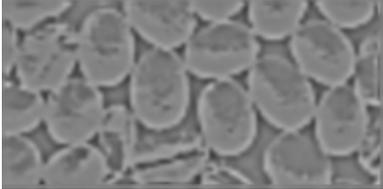 | 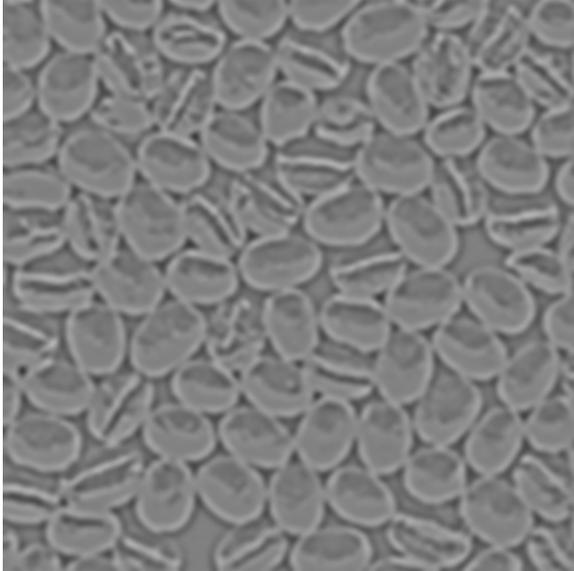 |
|---|---|
| $T\otimes = E(T,F) - B(E(T,F))$ | $S\otimes = E(E^{-1}(S,F)) - B(E(E^{-1}(S,F)))$ |

5.) Divide the enlarged target into an overlapping grid of square patches $\{t\otimes_1 ... t\otimes_n\} \in T\otimes$.

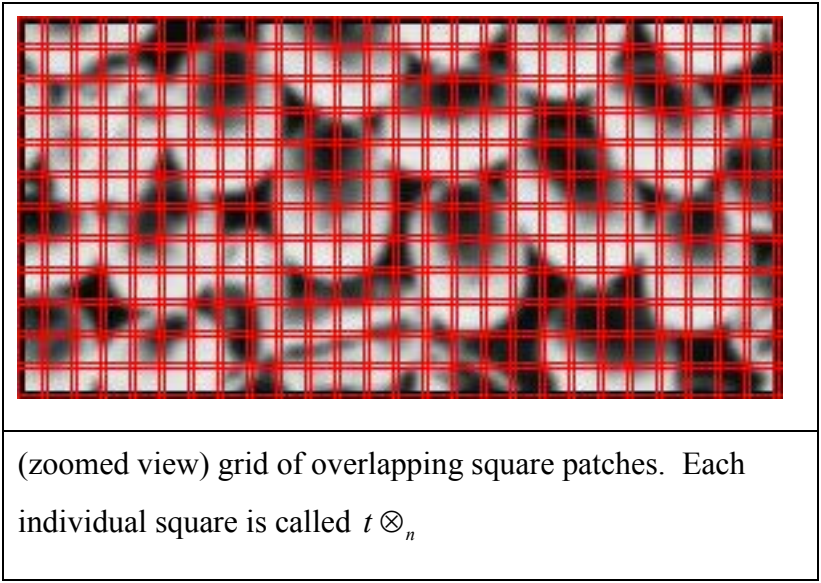

(zoomed view) grid of overlapping square patches. Each individual square is called $t\otimes_n$

For all n,

>Search for the closest match $s\otimes_{tn}$ to each patch $\{s\otimes_1 ... s\otimes_m\} \in S\otimes$ in the reenlarged source: $s_{tn} \in \{s\otimes_1 ... s\otimes_m\} | \forall j \in \{1..m\} \min(\|t\otimes_n - s\otimes_j\|)$. Distance is a weighted sum of two operands:
>
>>1. *Patch edge similarity:* The L1 distance between the luminosity of each pixel in $T\otimes$ and the corresponding pixel in $S\otimes$.
>>
>>2. *Patch luminosity similarity:* The L1 distance between the luminosity of each pixel in $E(T,F)$ and the corresponding pixel in $E(E^{-1}(S,F))$. (Sometimes the difference between pixels in both the A and B channels of LAB space were also used to provide clearer differentiation between the textures.)
>
>Call this patch the *found patch* $s\otimes_{tn}$.

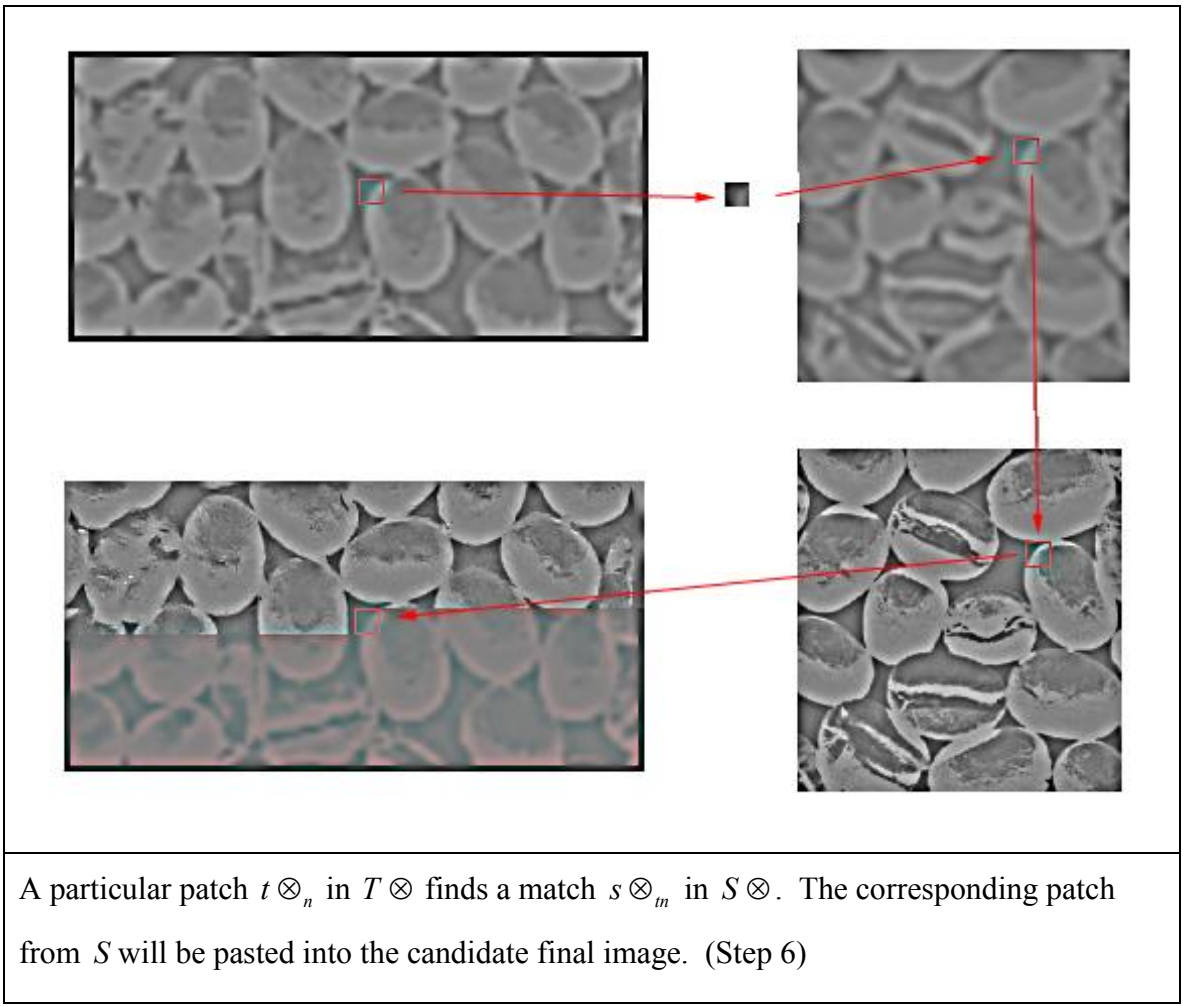

A particular patch $t\otimes_n$ in $T\otimes$ finds a match $s\otimes_{tn}$ in $S\otimes$. The corresponding patch from $S$ will be pasted into the candidate final image. (Step 6)

The search is done using a k-d tree assembled from the sample patches. We used the ANN (Approximate Nearest Neighbor) library to speed up the searching. This reduced the time to search by 3 orders of magnitude from brute force search.

6.) Place the source patch $s_{tn}$ that corresponds to the found patch $s\otimes_{tn}$ into the reconstructed image. In this way the comparison is done between images with medium frequency information, but the reconstruction is done with patches that contain high frequency information.

7.) Using the graph mincut algorithm, find where to cut the boundaries of the patch so that it fits well at the edges with neighboring patches.

| |
|---|
| The mincut algorithm finds an irregular path to cut along so that the difference between the two overlapping patches along the cut is minimized. |

8.) Add the blurred version of the enlarged target to the reconstructed image. This restores the low frequency information. Call this the *candidate final image*.
$C = B(E(T)) + R$

9.) We want to enforce the constraint that the reduced version of our final image looks exactly like the original target image. In order to do this we use a simple form of back-propagation, as follows:

1.) Reduce the candidate final image by the enlargement factor. $E^{-1}(C)$

2.) Find the difference between the reduced image and the target image.
$D = E^{-1}(C) - T$

3.) Enlarge the difference and add it to the candidate final image. $C + E(D)$

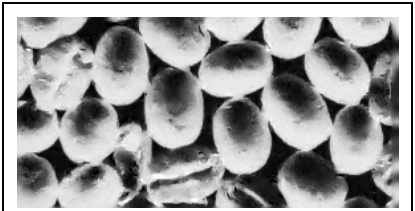

Final image $C + E(D)$

**Potential uses and limitations**

As a superresolution technique, this method works best on images that are made up of well-represented and clearly differentiated textures. Some images that might fall into this category are satellite images (with regions of farmland, water, and forest), medical scans (of a few distinct tissues), or text (in a known set of fonts). The reconstruction may not be correct in all details, but instead simply contains plausible details. This limits its use in cases where it is critical that no false detail information is acceptable.

| 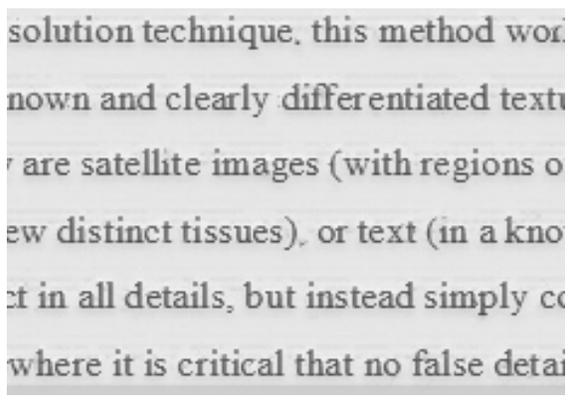 | 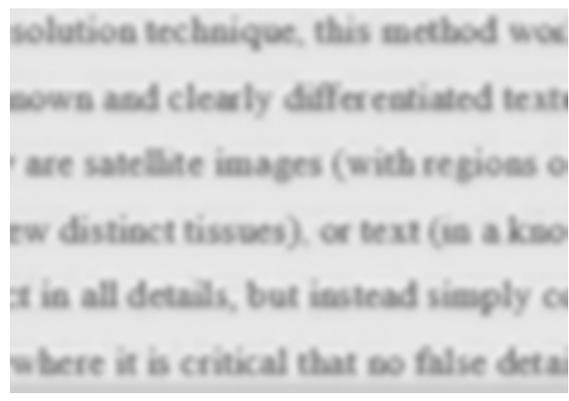 |
|---|---|
| text enlarged 2x | enlarged using bicubic |
| 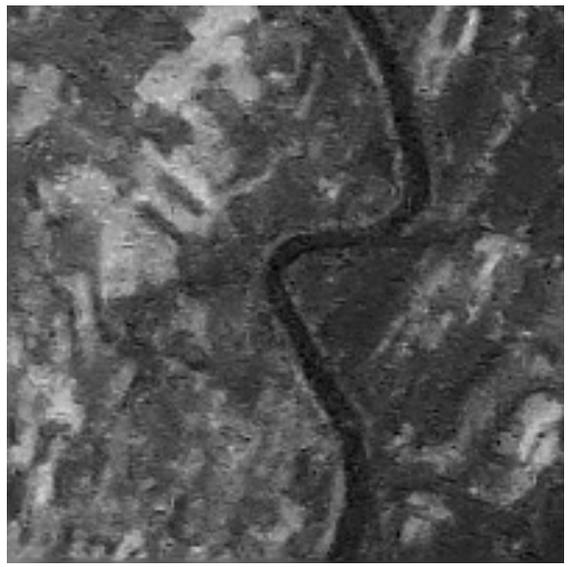 | 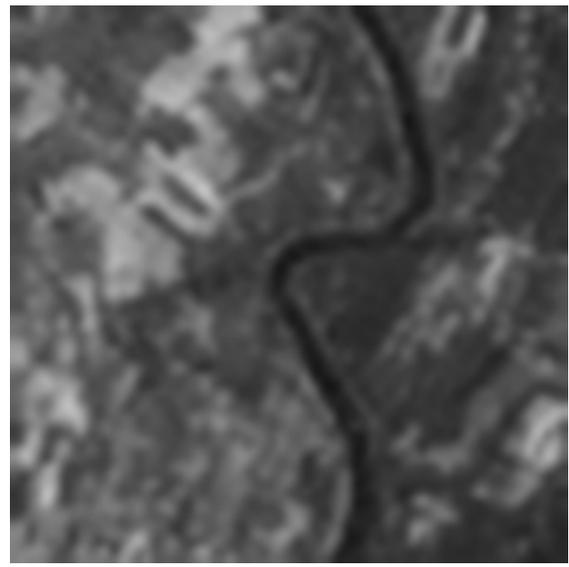 |
| satellite photo enlarged 3x | enlarged 3x using bicubic |

Many natural images are not composed of regions of simple textures. To have a good chance of accurately reconstructing a face in a generic photograph, for example, one would need samples of many faces from many angles and many lighting conditions. If the algorithm is given a small assortment of natural images as samples and asked to reconstruct another natural image, it will only be able to do a good job on features that are common to nearly all scenes. The main effect is to sharpen edges that are blurry in the bicubic-enlarged image, and adds some rough texture to overly smooth areas. The amount of processing time and memory that would be necessary to be able to enlarge general natural images with significantly better results than the best adaptive interpolation methods will likely not be available for many years.

The nature of the reconstruction and matching algorithm are entirely independent of the fact that the images being used for matching are low-resolution versions of the images used for reconstruction. In other words, with very minor modifications to the setup steps, this same program can be used as for texture transfer, colorization, switching modalities, artistic filters, or image analogies applications. Because graphcut synthesis is generally better than pixel-based synthesis, the results will typically be better than the original image analogies program.

**Background of the Problem**

Single-image superresolution (also known as expansion, zooming, resolution enhancement, or detail enhancement) is the process of reconstructing a high resolution image from two sources of information: a low resolution version of the same image, and our knowledge of the general class of images from which the image is taken. It is called "single-image" to differentiate it from the more common use of the term superresolution-- taking multiple images of the same scene (such as frames of a video) and combining the information to create a higher-resolution image.

The phrase comes from [13] a main source for many of the ideas in the field. But the overall problem is a very old one, under the name of image data interpolation. Each type of interpolation, from the simplest (nearest neighbor) to the most sophisticated, uses a model of the world in order to fill in the missing data between known pixels.

| Interpolation method | Image model | Artifacts include... |
| --- | --- | --- |
| Nearest neighbor | Nearby pixels tend to be the same color (intensity tends to be constant) | "Jaggies" (stairstep edges) |
| Bilinear | Smooth gradations of color are likely (The first partial derivatives of intensity tend to be constant) | Blurriness (lack of high frequency information) |

| Bicubic | Fitting image data with polynomial splines<br><br>(The second partial derivatives of intensity tend to be constant) | Blurriness, ringing at edges |
|---|---|---|
| Adaptive (NEDI, etc...) | Strong edges separate interpolation regions | Fails to reconstruct texture (The only high frequency information is at strong edges) |
| Fractal | Image details are similar to larger image features | Inappropriate details in texture (High frequency information is sampled from the wrong space) |
| Patch-based | Explicit example images | poor coherence between neighboring patches, insufficient samples |

The problem of filling in missing information is underconstrained. Assuming an 8-bit color space in the luminosity channel, an enlargement of 400% (2x in each direction) means filling in 3 unknown pixels for each known pixel, with $256^3$=16 million possibilities. However, the probability is very sharply peaked near the known values for a surprisingly large class of images. Very few pixels are more than 2 brightness values away in either direction. This reduces the space of possibilities to just $5^3$=125 for each known pixel. For this reason, even "nearest neighbor" interpolation is much better than choosing random colors to fill in the gaps between known pixels, because pixels tend to be of similar intensity to their neighbors. Bilinear interpolation of pixel values gives better results than nearest neighbor because the world contains many smooth gradations of color and few square patches of color that happen to be aligned with the pixel grid.

Bicubic interpolation is better than bilinear for most natural images, but already exhibits some artifacts at sharp edges that make higher-order polynomial interpolation less useful. (These are the result of intensity curves overshooting the data points.) Adaptive interpolation methods maintain sharp edges by interpolating along sharp image gradients, but not across them. At best, all these interpolation methods simply preserve the high frequency information that is already present in a low-resolution image. To get more realistic results or higher levels of resolution, it is necessary to introduce new high frequency information from some other source.

There are three places the new information could come from. One is generating the details algorithmically. The problem with this approach is a lack of a comprehensive theory of what kind of detail to generate. On the assumption that any kind of information will do, some enlargement software (such as S-Spline Pro) will simply add high frequency noise to the interpolated image. This tends to improve the perceived crispness of the image (how much it appears to be in focus) up to a point, but is clearly not an optimal solution for creating realistic detail. More complex generative models can be handcrafted, but these tend to have a limited domain of usefulness.

A second possibility is to use details of the image itself at the smaller scale. This is known as fractal interpolation, and is used in the context of fractal image compression. This is unlikely to work particularly well except on images that happen to be fractals, such as rocks, clouds, or cauliflower. Even then, small images may not contain enough samples to provide realistic details.

The third possibility is to introduce information from another source. In the last five years, researchers have begun to explore methods of superresolution that explicitly copy details from high resolution images. These were developed in the light of successful texture synthesis and texture inpainting algorithms. This approach has clear advantages over fractal interpolation, because the size of source images can be much larger than the original image. With fractal interpolation, we know that samples will be at the wrong scale, and hope that it won't matter too much. But using other images we have the possibility, at least, of finding details of a similar texture at a similar scale.

All of the texture synthesis super-resolution methods work in a similar way, but differ in ways to improve the quality and speed of the reconstruction. There are two main differences between this program and previous methods.

First, this program explicitly tries to choose patches that not only are of a similar luminosity at each point, but also come from an area with a similar overall texture. For example, in a forest scene, leaves should be reconstructed from samples of other leaves, while bark is reconstructed from samples of other bark.

Second, most texture synthesis superresolution methods use fairly simple techniques to reduce the impression that the reconstructed image has been assembled from independent patches. For example, neighboring rectangular patches may be overlapped and the overlapping pixels averaged. In contrast, this program uses graphcut texture synthesis, which adapts the shape of a patch to minimize artifacts.

# Previous Work

## Previous Work: Interpolation

Linear interpolation has been used to fill in gaps in tables since at least the time of the Babylonian Empire. Techniques equivalent to cubic spline interpolation were used by the ancient Greeks for astronomical tables [1]. Much research has been done on using adaptive techniques that use differently weighted interpolations depending on the direction of edges. For example, see [2] or [3].

## Previous Work: Texture Synthesis

Texture synthesis is the problem of generating a large area of texture from a small sample of that texture. It is very similar to the problem we are trying to solve, though simpler because it has fewer constraints and only deals with a single texture at a time (though often what is synthesized is composed of multiple independent elements which themselves would be considered textures.) [4]

### Pixel-based:

Pixel-based texture synthesis techniques create a new image one pixel at a time. Efros and Leung proposed assuming a Markov random field and estimating the conditional distribution of a pixel given the neighbors that have all ready been synthesized by sampling from the source texture [5].

The next year Wei and Levoy suggested that the algorithm could be made much faster by using tree structured vector quantization to perform the search instead of brute force. [6]

Ashikhmin noted that neighboring pixels usually met the statistical requirements and were very easy to find, speeding up the algorithm further. His synthesis technique built the image out of irregular patches taken from the source image [7].

In Image Analogies [8] Hertzmann et. al. used pixel-based texture synthesis techniques to build up their images.

**Patch based:**

There are two main problems with pixel-based texture synthesis. First, it is extremely slow. Second, scan-line fill order can lead to ugly artifacts that grow diagonally downward at a 45-degree angle. Both of these problems can be fixed by observing that pixels that are adjacent in the source texture are probably correct choices for reconstructing adjacent pixels in the synthesized texture. If too large a region is used, it will be obvious that it is copying directly from the source image, but with patches between 2 and 20 pixels this is unlikely to be a problem. The only remaining problem is how to ensure that neighboring patches fit together well.

The first papers that introduced this method [9] [10] used matching constraints between overlapping patches in their search, then simply averaged (feathered) the overlapping portions of the patches.
Feathering is not the best solution because it tends to destroy high frequency details of the patches. In Graphcut Textures, [11] Kwatra et. al. find a cut between overlapping patches that minimizes the visual discontinuity between the two.
In general, the best solution to combine two images is to blend the low frequency information and to perform a minimum cut between the high-frequency portions of the image (known as multi-band blending). Since this program only synthesizes the high frequency details, however, graphcut is the best solution for preserving texture details.
Hybrid Texture synthesis [12] is another promising approach to the problem. It uses overlapping patches, but wherever the overlapping portions are sufficiently different, it goes back to pixel-based texture synthesis techniques. Since more of the neighborhood is already filled in, and only small portions are overlapping, this produces excellent results. It would be a good area to look into for someone wishing to extend this superresolution technique.

**Previous Work: Patch Based Superresolution**

The work by Freeman in 2000 [13] was the seminal paper in this field. The program described in this paper is simply an incremental advancement on the method described here.

Freeman was primarily interested in learning to synthesize general natural images from other general natural images. Because of this, his methods perform well at reconstructing edges, but is not really designed to reproduce textures well. He uses feathering between overlapping patches.

The paper "Super-Resolution through Neighbor Embedding" [14] suggest using LLE (locally linear embedding) to blend multiple patches together. This is another idea that improves the results on edges and corners, but loses texture information (see the paragraph on LLE in the "Discussion of Features" section of this paper.)

Research at Microsoft [15] suggests using a different basis that is explicitly tailored towards reconstructing edges, under the assumption that this is all that can be learned well.

One of Ashikhmin's last papers before he gave up the whole field of graphics in disgust was Fast Texture Transfer [16]. This uses his pixel-based texture synthesis technique described above to do texture transfer and superresolution.

The most similar technique to the one described in this paper is an as yet unpublished paper by Ganesh Ramanarayanan and Kavita Bala on constrained graphcut texture synthesis. A poster on it can be seen here:

http://www.cs.cornell.edu/~graman/projects/cgs/.

A few researchers have begun to apply similar techniques in restricted domains with higher quality results. Two examples are camera oriented human faces (http://www.cs.rutgers.edu/~mlittman/courses/ml03/iCML03/talks/plenary/huang.pdf) and 3-D volumes [18].

## Discussion of Features

### Graphcut Synthesis

Much attention has been given to the problem of what to do with the edges between patches in texture synthesis. The simplest approach is to average the overlapping pixels (feathering). This leads to visible blurring along the edges. Feathering works well for combining low frequency visual information, but to preserve the high frequency information it is better to try to find the cut that minimizes visual discrepancy between the two sides of the cut. The discrepancy is expressed as an energy term, and we try to find the minimum energy.

The energy equation used for defining the cut in this program is

$$E = |a_{x,y} - b_{x,y}| + |a_{x,y+1} - b_{x,y+1}| \text{ and } E = |a_{x,y} - b_{x,y}| + |a_{x+1,y} - b_{x+1,y}|$$

Where:

$E$ is energy

$a$ is the previously laid down patches

$b$ is the new patch

$x$ and $y$ are the pixel coordinates

Graphcut texture synthesis [11] uses the min-cut/max-flow graph segmentation algorithm to find the optimal cut between two texture patches.

This program uses overlapping square patches, but the overlap region is cut by graphcut.

### Approximate Nearest Neighbor Search (ANN)

Even for small (200 x 200) source and target images, it can take hours to search for the optimal patches using brute force. Instead, this program creates a k-d tree from the source image patches to speed up the search so that it can be completed in less than a minute for small images.

The ANN library by David Mount and Sunil Arya was used to create and search the tree. (This same library was used in [8].) The search is "approximate" because it finds either the closest match among all the source patches, or a patch that is within epsilon of a perfect match using the specified metric. For all the examples in this paper epsilon was set at $\varepsilon = .5w^2$ where $w$ is the width of the patch and the L1 (Manhattan) metric was used to compare patches.

The library routines to access memory were modified slightly so that it takes only a small constant times the size of the original image to store the search tree, at the cost of a slight increase in search time.

**Earth Mover's Distance Metric (EMD)**

The program tries to find the closest match between the low-resolution source and the low-resolution target. "Closeness" is defined in terms of a particular metric. Three potentially useful metrics for comparing patches are the Minkowski L2 norm (Euclidean distance) the L1 norm (Manhattan distance) and the Earth Mover's Distance. Ultimately, the metrics themselves must be judged by how well they correspond to human perception of similarity between the patches. The L2 norm seemed too sensitive to a large variation in a single pixel compared with the results using the L1 norm. The Earth Mover's distance was designed to be closer to human perceptual distance. Here is an intuitive explanation of the EMD:

Imagine the patch as an arrangement of stacks of building blocks, where each stack is as high the luminosity value at the corresponding pixel. Moving one block, one pixel at a time, what is the minimum number of moves it would take to transform the sample into the target?

Consider the case of black patches with a single white pixel. The L1 distance metric would show all of these patches as being the same distance from each other, no matter where the white spot was in the patch, except for the case where it happens to fall on exactly the same pixel. The Earth Mover's Distance metric correctly matches the patches with a nearby white pixel as more similar than the ones where the white pixel is far away. For small patch sizes and with the blurry patches we are trying to match the difference between the results of EMD and L1 are very subtle, and the EMD is much slower to compute. For larger patch sizes, the EMD may be necessary to accurately match patches.

**Locally Linear Embedding (LLE)**

Locally linear embedding assumes that the mapping between the space of low-resolution patches and the space of corresponding high-resolution patches is locally linear. Instead of using the closest match for a patch, we can take the closest n patches and find an optimally weighted average of them.

For each patch

There are a few problems with this approach. One is oversmoothing: assuming all the original patches contain noise (or any kind of very high-frequency stochastic texture) averaging them together will reduce this noise. Sometimes noise reduction is a benefit, but in this case the noise is part of the texture we are trying to preserve.

A second problem is that the resulting averaged patches do not come directly from the source distribution. The optimal solution may give some of the patches in the solution negative weights. This can create patches whose low-resolution version is a close match to the target, but whose high-resolution version doesn't resemble anything in the original high-resolution source. In this case, the assumption that the embedding is locally linear has proven incorrect.

Finally, it is difficult to use the L1 or EMD metric with this approach, because unlike the L2 metric simply solving an appropriate system of linear equations is insufficient.

For these reasons, LLE was not used in the final version of the program.

**Backpropagation**

We want to enforce the constraint that the reduced resolution version of our final image looks exactly like the original target image. In order to do this we use a form of back-propagation, as follows:

    Reduce the candidate final image by the enlargement factor.
    Find the difference between the reduced image and the target image.
    Subtract a fraction of the difference image from the previous candidate final image.
    Repeat multiple times until convergence.

**Luminosity Equalization**

This was suggested by Freeman. A scene under different lighting conditions or captured by a different camera may have very different luminosity values, but the features to be learned are the same. Subtracting out the low frequencies of the image (known as a high pass filter) eliminates these differences, leaving only the high frequency features to be learned.

For reconstructing natural images from a small number of samples, this technique was very useful. However, the overall luminosity of a patch does tell something about the texture it came from. By throwing away this information we run the risk of using patches from the wrong texture to do the reconstruction. In this way it performs a function very similar to the information from the A and B channels which we explicitly included to improve the texture sensitivity. In the program the weight of this luminosity is set as a tunable parameter.

It is also possible to normalize for contrast as well as brightness. Rather than subtracting the low-passed image, one can divide by it, adding a small constant to prevent division by zero.

**Rotations and Reflections**

The k-d tree search is very fast, but takes a lot of memory. The same structure can be searched by transforming the input vector, searching, and doing the inverse transformation on the result vector. I used horizontal and vertical reflection and 180 degree rotation. It wouldn't be difficult to do other rotation angles, too. But this slows down the search.

**Inter-Patch Smoothness Constraint**

This was suggested in [13]. Where the patches overlap, I search through the close matches and find the one that best matches the overlapped pixels. This makes unnatural jumps between patches less likely.

**Additional Results**

Top: True image composed of four textures. Second row: Enlarged using block interpolation. Third row: Result of algorithm using samples from all four textures. The matching is good enough that patches come from the correct texture. Bottom: Using only the sample for the texture in the lower left-hand corner. Other types of textures are not reconstructed well, since there is no good source for them.

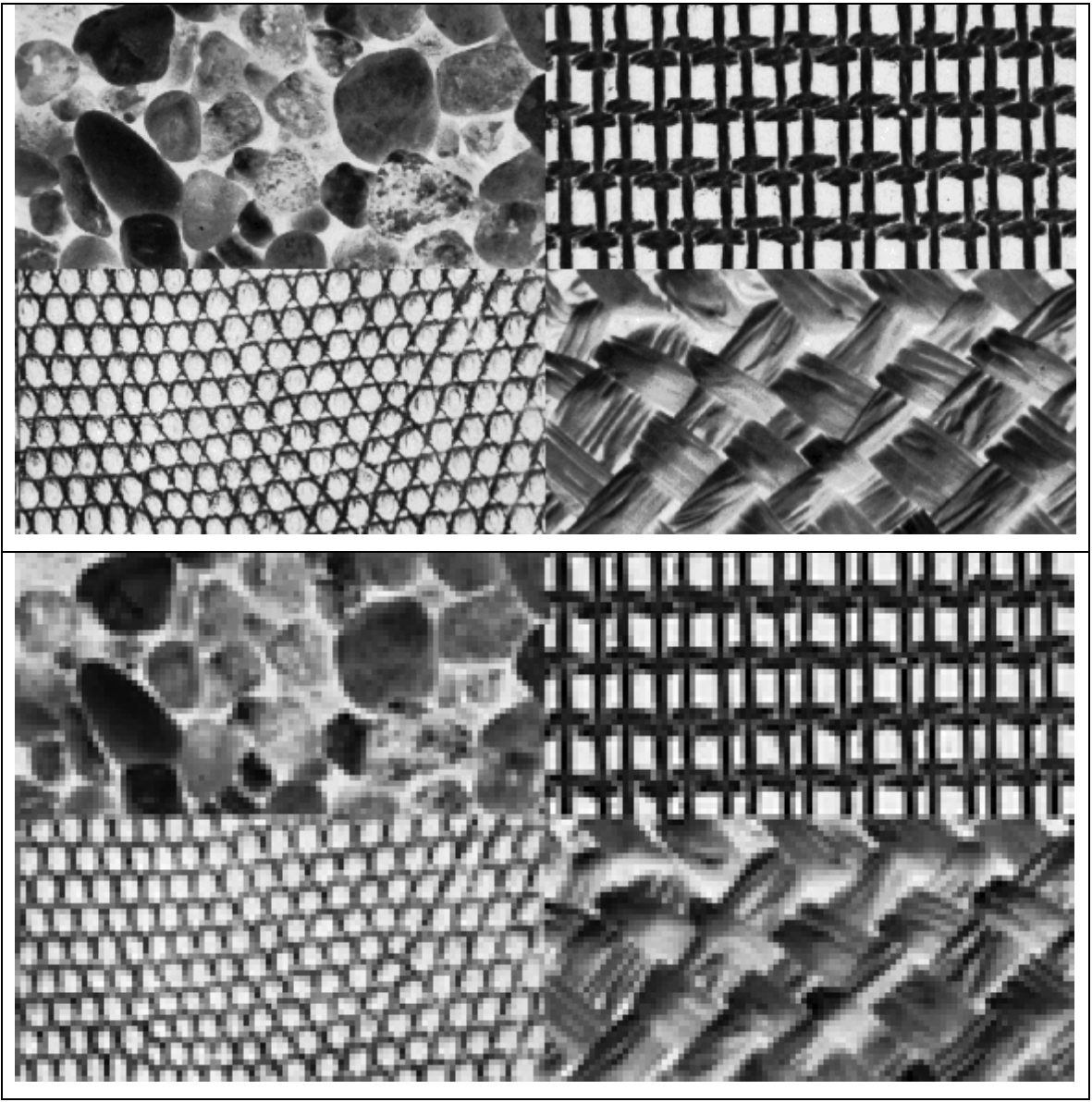

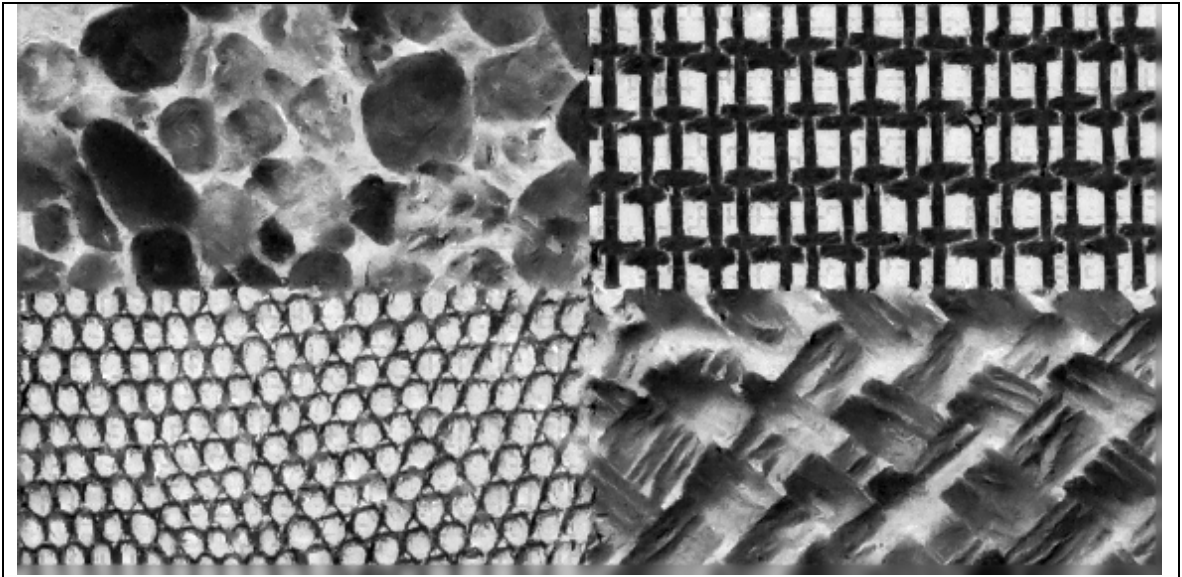
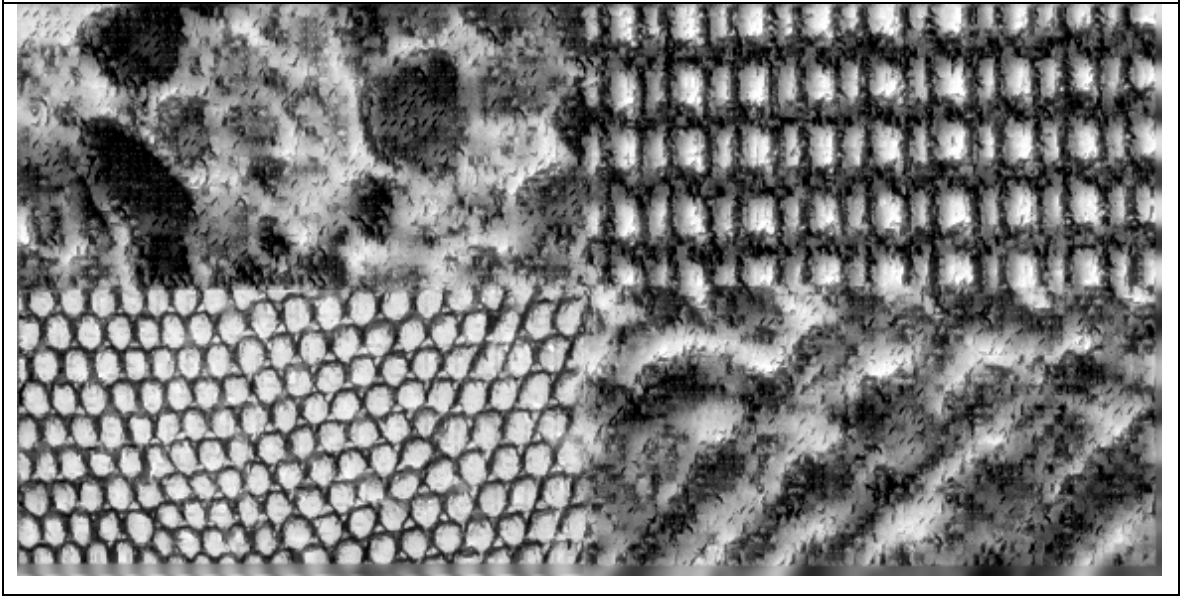

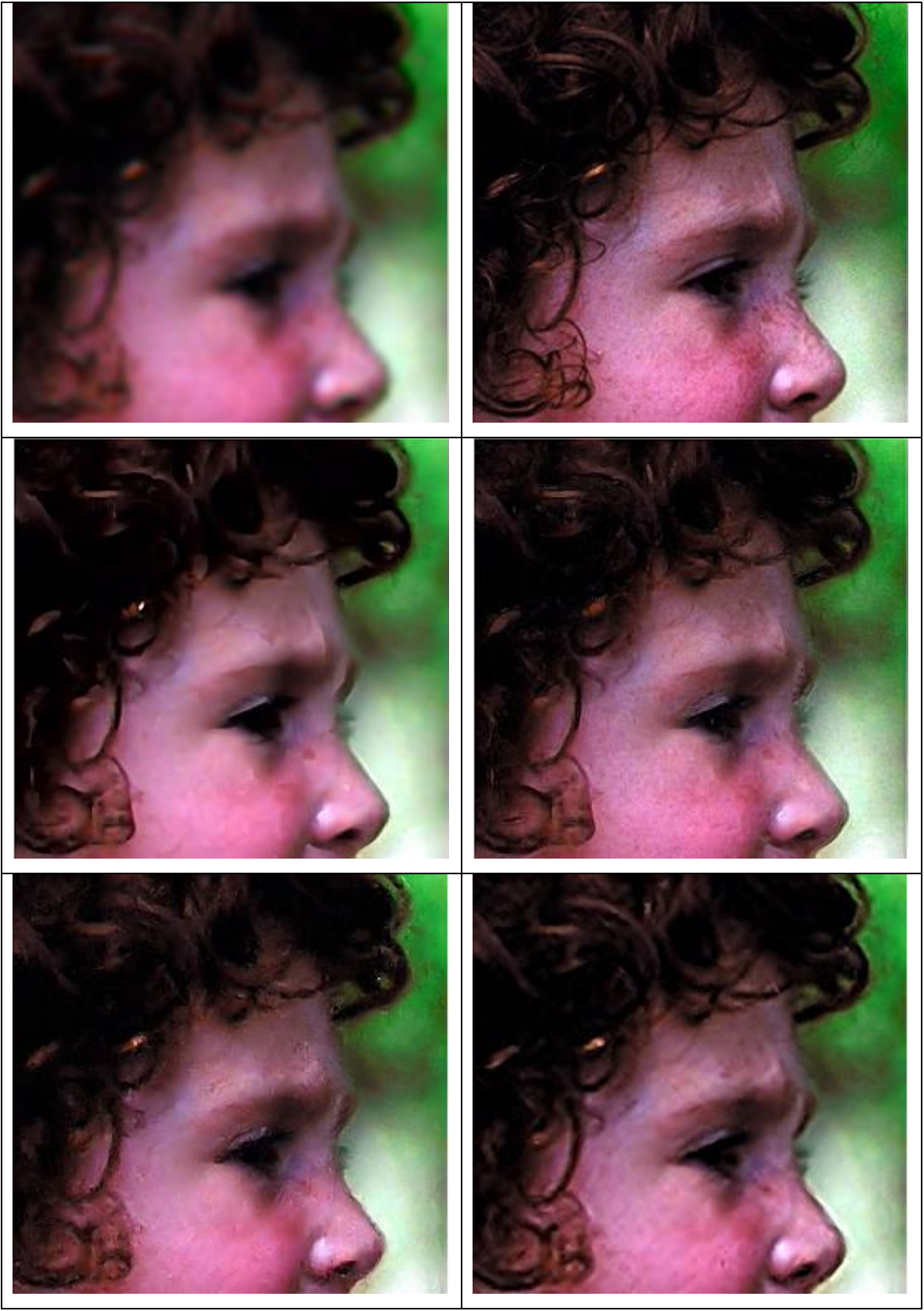

Image enlarged by 1600%, from 70 x 70 to 280 x 280. Upper left: bicubic interpolation. Upper right: full resolution. Center left: Freeman's results. Center right: results using rotated image as source texture. Lower left: using generic image as source. Lower right: using generic image as source, enlarging by 2x and then enlarging the result by 2x.

**Citations**